\documentclass{article}

\usepackage{arxiv}

\usepackage[utf8]{inputenc} 
\usepackage[T1]{fontenc}    
\usepackage{hyperref}       
\usepackage{url}            
\usepackage{booktabs}       
\usepackage{amsfonts}       
\usepackage{nicefrac}       
\usepackage{microtype}      
\usepackage{lipsum}		
\usepackage{graphicx}
\usepackage{doi}

\usepackage{physics}
\usepackage{bm}

\usepackage[sorting=none]{biblatex}
\usepackage[normalem]{ulem}
\usepackage{xcolor}
\newcommand{\Add}[1]{\textcolor{black}{#1}} 
\newcommand{\Del}[1]{\if0{#1}\fi} 
\newcommand{\Rep}[2]{\Del{#1} \Add{#2}} 

\newcommand*{\T}{\mathsf{T}}
\newcommand*{\diff}{\mathrm{d}}

\newcommand*{\vect}[1]{\bm{#1}}

\newcommand*{\crm}{\mathrm{c}}

\newcommand*{\hrm}{\mathrm{h}}

\newcommand*{\nrm}{\mathrm{n}}

\newcommand*{\srm}{\mathrm{s}}

\newcommand*{\Irm}{\mathrm{I}}

\newcommand*{\R}{\mathbb{R}} 

\newcommand*{\targ}{\mathrm{targ}}

\title{Optimization of body configuration and joint-driven attitude stabilization for transformable spacecrafts under solar radiation pressure}

\author{ \href{https://orcid.org/0000-0001-7710-9504}{\includegraphics[scale=0.06]{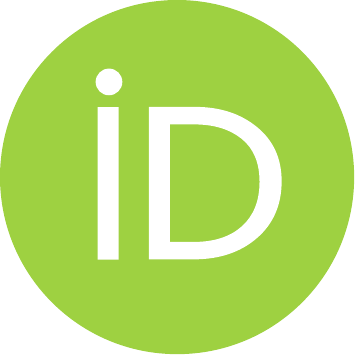}\hspace{1mm}Yuki ~Kubo} \\
	Institute of Space and Astronautical Science\\
	Japan Aerospace Exploration Agency\\
	Sagamihara, 252-5210, Japan \\
	\texttt{kubo.yuki@jaxa.jp} \\
	\And
	\href{https://orcid.org/0000-0001-5739-3982}
 { \includegraphics[scale=0.06]{orcid.pdf}\hspace{1mm}
    Toshihiro Chujo} \\
	Department of Mechanical Engineering\\
	Tokyo Institute of Technology\\
	Ookayama, 152-8550, Japan \\
	\texttt{chujo.t.aa@m.titech.ac.jp} \\
}

\renewcommand{\shorttitle}{\textit{arXiv} Template}

\hypersetup{
pdftitle={arxiv version},
pdfsubject={q-bio.NC, q-bio.QM},
pdfauthor={Yuki Kubo},
pdfkeywords={First keyword, Second keyword, More},
}

\begin{document}
\maketitle

\begin{abstract}
A solar sail is one of the most promising space exploration system because of its theoretically infinite specific impulse using solar radiation pressure (SRP). Recently, some researchers proposed "transformable spacecrafts" that can actively reconfigure their body configurations with actuatable joints. The transformable spacecrafts are expected to greatly enhance orbit and attitude control capability due to its high redundancy in control degree of freedom if they are used \Rep{as}{like} solar sails. However, its large number of input poses difficulties in control, and therefore, previous researchers imposed strong constraints to limit its potential control capabilities.
This paper addresses novel attitude control techniques for the transformable spacecrafts under SRP. The authors have constructed two proposed methods; one of those is a joint angle optimization to acquire arbitrary SRP force and torque, and the other is a momentum damping control driven by joint angle actuation. 
Our proposed methods are formulated in general forms and applicable to any transformable \Rep{solar sail}{spacecraft} \Rep{that consists of flat and thin body components}{that has front faces that can dominantly receive SRP on each body}. Validity of the proposed methods are confirmed by numerical simulations. 
This paper contributes to making most of the high control redundancy of transformable \Rep{solar sails}{spacecrafts} without consuming any expendable propellants, which is expected to greatly enhance orbit and attitude control capability. 
\end{abstract}

\keywords{Solar sail \and Transformable spacecraft \and Attitude–joint coupled motion \and Attitude stabilization }

\section*{\normalsize Nomenclature}
\begin{center}
    \begin{tabular}{|l|l|}
        \hline
        CoM & center of mass \\
        \hline 
        SRP & solar radiation pressure \\
        \hline
        $\theta_k$ & angular displacement of the $k$-th actuatable joint (rad)\\
        \hline
        $\theta$ $\qty(=[\theta_1, \cdots, \theta_m]^\T)$ & joint angle vector (rad)\\
        \hline
        $\lambda_k$ & rotational axis of the $k$-th joint\\
        \hline
        $\omega$ & angular velocity of body $0$ (rad/s)\\
        \hline
        $R_X$ & absolute position of the CoM of body $X$ (m)\\
        \hline
        $r_{XY}$ \Add{$\qty(=R_X-R_Y)$} & relative position of \Add{the CoM of} $X$ with respect to \Add{the CoM of} $Y$ (m)\\
        \hline
        \Add{$r_{X}$ $\qty(=R_X-R_0)$} & \Add{relative position of the CoM of $X$ with respect to the CoM of body 0 (m)}\\
        \hline
        $h_\crm$ & angular momentum around the CoM of whole bodies (kg$\cdot$ m$^2$/s)\\
        \hline
        $I_X$ & moment of inertia of $X$ around the CoM of the $X$ (kg$\cdot$ m$^2$)\\
        \hline
        $m_X$ & total mass of $X$ (kg)\\
        \hline
        $M$ & generalized mass matrix \\
        \hline        
        $v$ & generalized velocity \\
        \hline
        $w$ & generalized acceleration \\
        \hline
        $d$ & second-order term of a generalized velocity\\
        \hline
        $\tau$ & generalized force\\
        \hline
        $U$ & identity matrix \\
        \hline
        $x^\times$ & skew symmetric matrix of a vector $x\in\R^3$\\
        \hline
        $(\cdot)_k$  & subscript which signifies the $k$-th body\\
        \hline
        $(\cdot)_\crm$ & subscript which signifies the whole bodies\\
        \hline
        $(\cdot)_{\hat{k}}$ & subscript which signifies the $k$-th outer group\\
        \hline
        $(\cdot)_{\hrm_k}$ & subscript which signifies the $k$-th joint\\
        \hline
        $\Add{(\cdot)_{\srm_{ij}}}$ & \Add{subscript which signifies the $j$-th surface of the $i$-th body}\\
        \hline
        $\phi$ \Add{$\qty(=[\phi_1, \phi_2, \phi_3]^\T)$} & Euler angles (rad)\\
        \hline
        $C$ & direction cosine matrix \\
        \hline
        $t$ & time (s)\\
        \hline
        $s$ & sun pointing vector \\
        \hline
        $n_i$ & normal vector of $i$-th body\\
        \hline
        $P$ & solar radiation pressure (SRP) (Pa)\\
        \hline
        $F$ & SRP force (N)\\
        \hline
        $T$ & SRP torque (N$\cdot$m)\\
        \hline
        $C_{\mathrm{spe}}, C_{\mathrm{dif}}$, $C_{\mathrm{abs}}$ & coefficient of optical properties \\
        \hline
        $Q$, $R$ & weight matrices for LQR feedback control \\
        \hline
    \end{tabular}
\end{center}

\section{Introduction}

\subsection{Background}
\Add{Solar radiation pressure (SRP) is a potential propulsion resource in inter-planetary space missions, and its active use for space flight has been investigated for a long time.}
\Add{In particular,} \Rep{Solar sailing}{a solar sail, which has large and thin membrane structure,} is promising as a propellant-free propulsion system using \Rep{solar radiation pressure (SRP)}{SRP}, and has been investigated by a lot of researchers \cite{sauer1976optimum, leipold1998solar,mcinnes1994solar,baoyin2006solar, ono2016generalized, tsuda2019classification, liu2015dynamic, liu2017solar}. 
A solar sail controls its orbit by maneuvering attitude with respect to the sun, and therefore its attitude control method has crucial impacts on its orbit control capability. Attitude control methods of solar sails are classified into major two types depending on their membrane deployment systems: 1) spin-type, 2) boom-type. The former type deploys its membrane by centrifugal force of spinning motion without any stiff structures, which was demonstrated by IKAROS \cite{mori2010first, tsuda2011flight}. The spin-type is preferable for large-size sail in terms of mass and besides, can naturally stabilize its attitude by rotational stiffness. However, it requires more time and energy to change its attitude and cannot slew its orientation flexibly and agilely.
In contrast, the boom-type uses stiff support structure to maintain the shape of its membrane and usually adopts three-axis attitude stabilization with actuators such as magnetorquers and reaction wheels \cite{spencer2021lightsail}. However, attitude stabilization of large membrane requires very large torque because its moment of inertia increases in proportional to $r^4$ where $r$ is a radius of the membrane.
Moreover, neither of both types can orient its spacecraft bus independently of the membrane, and thus limits permissible attitude of the sail due to orientation constraints imposed on bus components such as antennas, cameras, and thrusters. All of the above problems restricts orbit control capability of solar sails.\par

As promising solutions to these problems, some researchers have been proposing novel solar sailing techniques in which the shape of the membrane is dynamically and independently controlled. Takao \cite{takao2018active, takao2019optimal, takao2020self} has proposed active shape control of sail membrane for spin-type solar sails. The main concept is to excite resonant oscillation on the membrane surface synchronized with its spin frequency, so that the shape of the membrane is actively changed seen from inertial frame. This technique enables the orientation of membrane and a spacecraft bus to be independently controlled and contributes to enhancing orbit control capability. \Add{However, the nonlinear dynamics of the dynamically oscillated membrane causes a lot of difficulties in control. In addition, achievable orientation of the membrane is constrained by magnitude of excitable oscillation.} \par

Another promising solution is to add active joints to a \Rep{spacecraft}{solar sail}\Del{ to enhance redundancy of solar sailing} \Add{, which is referred to as a transformable solar sail}. The actuatable joints enable \Rep{body components of the spacecraft}{relative geometry of sail membranes} dynamically reconfigurable, and thus can contribute to satisfying \Del{multiple} orientation constraints of instruments while SRP torque and force are properly controlled to obtain desired attitude and orbit. Moreover, \Rep{the spacecraft with actuatable joints}{the transformable solar sail} can carry out propellant-free attitude reorientation enabled by nonholonomic properties of attitude equations of free-floating multi-bodies \cite{nakamura1990nonholonomic, papadopoulos1992path, murray1993nonholonomic, ohashi2018optimal, gong2022shape, kubo2022approximate, kubo2022nonholonomic}.
Chujo \cite{chujo2022propellant} formulated attitude motion of solar sails with variable-shape mechanisms in which large deformation of body components is properly handled without approximations. In addition, Chujo carried out stability analysis and propellant-free attitude maneuver for a specific spacecraft with four actuatable paddles. 
Another example is Abrishami and S. Gong's work \cite{abrishami2020optimized} that proposed a redundant control method for a solar sail with four actuatable petals equipped with reflectivity control devices (RCD). They solved a complex mixed-integer-continuous and nonlinear optimization problem by approximating SRP torque with a modified egg-shaped equation. 
\Add{However, a degree of freedom of actuatable joints is limited to 2--4 in most transformable solar sails because flexible membrane structure cannot mechanically hold heavy joint mechanisms on it. This mechanical constraint makes it difficult for transformable solar sails to simultaneously satisfy orientation constraints of multiple instruments.}
\Del{However, control strategy of these works and most other works largely depend on the simplicity (and in most case, symmetricity) of the solar sail model. Chujo's another work proposed a control strategy with highly redundant transformable solar sail, but it sacrificed most of the control degree of freedom to simplify the problem \cite{chujo2022orbit-attitude}.
Thus, attitude control strategy of solar sail with large number of active joints has not generally been derived.}
In addition, most formulations have not dealt with attitude-joint coupled dynamics. Chujo's work \cite{chujo2022propellant} and Abrishami's work \cite{abrishami2020optimized} have not considered dynamic joint actuation during solar sailing and H. Gong's work \cite{gong2022attitude} has not included effect of SRP torque in its attitude control. However, it is crucial to derive such a control law that properly handles attitude-joint coupled dynamics to make the most of transformable solar sails. \par
\Add{Recently, another novel and promising spacecraft, a transformable spacecraft is investigated by several researchers including the authors \cite{kubo2022preliminary, miyamoto2023orbit}. The transformable spacecraft is composed of multiple rigid body components connected with electrically actuatable joints and can reconfigure geometry of body components dynamically. Unlike the transformable solar sail, the transformable spacecraft can install a lot of actuatable joints on rigid body components. Thus, the highly redundant degree-of-freedom in control enables it to satisfy orientation constraints of multiple instruments, and besides, to carry out more agile nonholonomic attitude reorientation maneuver. 
However, control strategy of previous works largely depend on the simplicity (and in most case, symmetricity) of specific models of transformable spacecrafts. Chujo's another work proposed a control strategy with a highly redundant transformable spacecraft, but it sacrificed most of the control degree of freedom to simplify the problem of attitude and body configuration optimization under SRP \cite{chujo2022orbit-attitude}. Thus, attitude control strategy of transformable spacecrafts with large number of active joints has not generally been derived. Moreover, attitude-joint coupled dynamics have not been solved to plan attitude maneuver as is the case with the transformable solar sails \cite{chujo2022propellant,abrishami2020optimized,gong2022attitude}.}

\subsection{Contributions}
In this paper, we propose an attitude-joint coupled control method that can be applicable to arbitrary \Rep{transformable solar sails}{transformable spacecrafts} under SRP exposure. 
Our contribution is divided into two novel techniques: 1) joint angle optimization to satisfy arbitrary SRP force and torque constraints (\textit{SRP-based joint angle optimization}), 2) angular momentum damping control by attitude-joint coupled control. The former method contributes to obtaining neutrally stable equilibrium attitude and body configuration under SRP field to acquire arbitrary SRP acceleration. Unlike the previous method \cite{chujo2022orbit-attitude}, our technique can handle general body configurations without restricting the control degree of freedom, and thus greatly enhances solar sailing capability of transformable spacecrafts. The latter technique, the angular momentum damping control, contributes to asymptotically stabilizing the equilibrium attitude obtained by the above joint angle optimization technique. We formulated a generalized attitude motion that can handle attitude-joint coupled motion to obtain the control law.
This technique can also be applicable to arbitrary transformable spacecraft and does not require any other additional actuators such as reaction wheels and reflectivity control devices.

\subsection{Organization of the paper}
Section \ref{sec:prelim} provides preliminary formulations to prepare for the following sections. Section \ref{sec:methods} formulates the two novel proposed techniques: joint angle optimization and angular momentum damping control. Section \ref{sec:sim} describes setups of numerical simulations and shows their result. Section \ref{sec:concl} summarizes the contents of the present paper and provides concluding remarks.

\section{Preliminary formulations} \label{sec:prelim}
This section provides some preliminary formulations for the following sections. 
Nomenclature used in the following formulations is listed in the beginning of this paper.
Throughout this paper, we clearly distinguish a vector and its component; a mathematical symbol with bold font stands for a vector entity, and that with a normal font stands for a component matrix of the vector expressed with a certain set of reference basis vectors. For example, dot-product and cross-product operations are defined for vectors as $(\vect{x}\cdot\vect{y})$ and $\vect{x}\times\vect{y}$ respectively while the same operations are expressed with $(x^\T y)$ and $(x^\times y)$ for vector components.
In addition, all vector components are expressed in body frame unless their reference frame is explicitly indicated. 

\subsection{Attitude dynamics of free-flying multi bodies}
\begin{figure}[hbt!]
	\centering\includegraphics[width=4.0in]{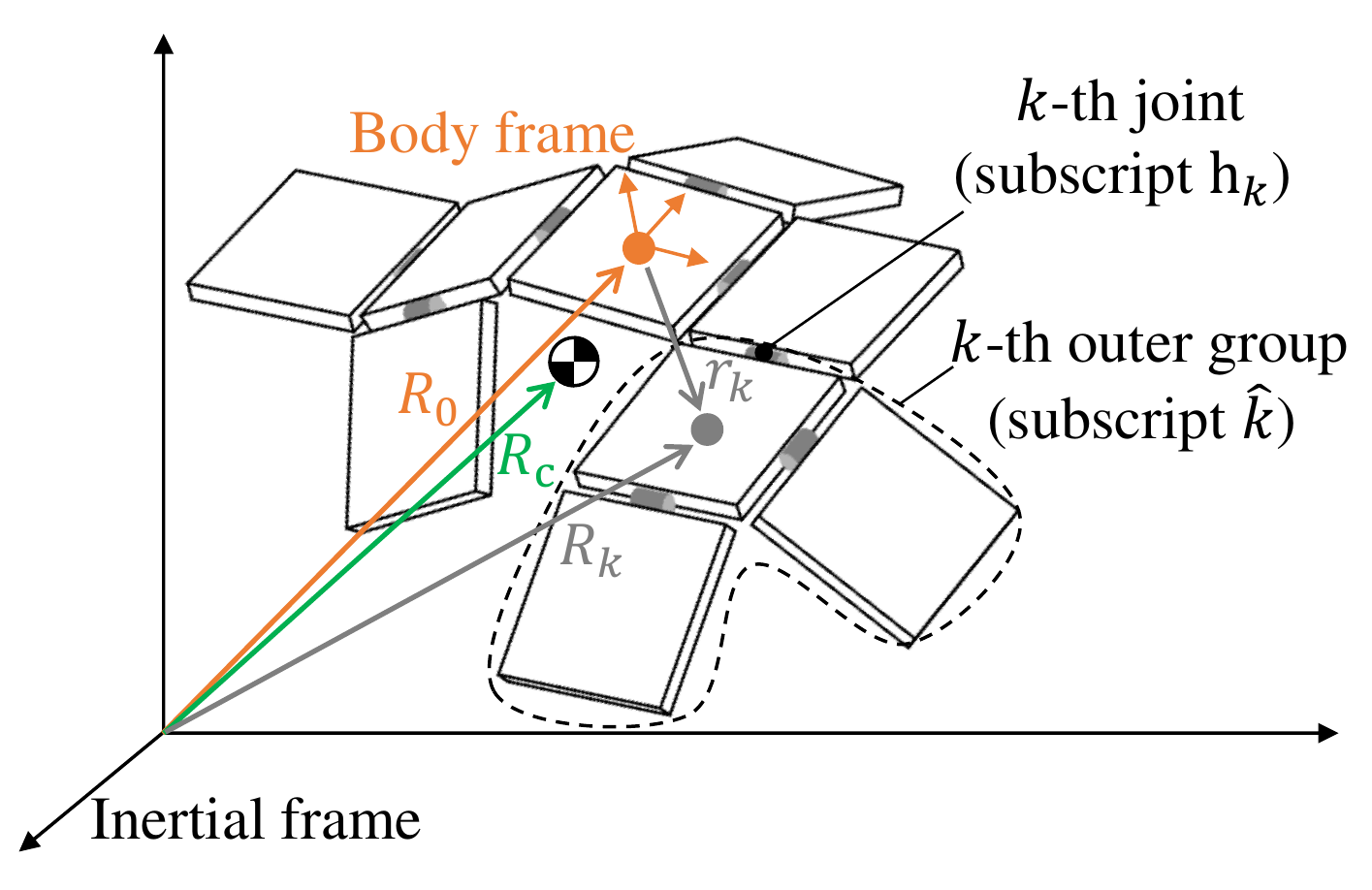}
	\caption{Nomenclature for body components and their positions}
	\label{fig:multi_rigid}
\end{figure}

The formulation of this section largely follows the author's previous work \cite{kubo2022simultaneous}.
The nomenclature for body components and their relationship are shown in Fig. \ref{fig:multi_rigid}. 
\Add{The characters defined in this figure can be interpreted with the nomenclature list at the beginning of this paper. $R_k$, $R_0$ and $R_\crm$ represent the absolute position of the CoM of $k$-th body, $0$-th body and whole bodies respectively, whereas $r_k$ expresses the relative position of the CoM of the $k$-th body with respect to the CoM of body 0.}
In this representation, an entire spacecraft is divided into a main body that is assigned an index $k=0$, and some other branches growing from the main body. Here, for convenience of formulation, we assume that each branch has no loop structure. This assumption assures serial index assignment from the main body (innermost, root) to the tip of each branch (outermost, tips). Body indices in each branch are allotted sequentially from an inner body to an adjacent outer body. The hinge joint component connecting the $k$-th body and its inside neighbor is labeled a $k$-th joint, and a set of all bodies outside of the $k$-th joint are labeled $\hat{k}$, which is referred to as $k$-th outer group. Note that all indices in the outer group $\hat{k}$ are larger than $k$ from the above definitions.
This $\hat{k}$ grouping contributes to simple formulations because all bodies in the $k$-th outer group move in the same manner when the $k$-th joint is actuated. The rotational degree of freedom of each hinge joint is set to be 1, and the rotational angle of the $k$-th joint is described as $\theta_k$. Any joint rotation can be expressed by a combination of 1-dimensional rotations, and hence this assumption does not lose generality. The body frame is attached to the main body ($k=0$), and its origin is at the center of mass (CoM) of the main body. Thus, the attitude of the entire multi body system is represented by the attitude of the main body hereafter. Note that the position of the CoM for the entire spacecraft is not fixed in the body-fixed coordinate as the spacecraft changes its body configuration.\par
We use the Kane's formulation in the present paper, which generally contributes to rapid numerical computation \cite{kane1983use,kane1985dynamics}. A general form of the dynamics equation can be described as follows:
\begin{equation}\label{eq:gen_eom}
	\begin{bmatrix}
		M_{vv} & M_{v\omega} & M_{v\theta} \\
		M_{\omega v} & M_{\omega\omega} & M_{\omega\theta} \\
		M_{\theta v} & M_{\theta\omega} & M_{\theta\theta}
	\end{bmatrix}
	\begin{bmatrix}
		w_v \\
		w_\omega \\
		w_\theta
	\end{bmatrix}
	+
	\begin{bmatrix}
		d_v \\
		d_\omega \\
		d_\theta
	\end{bmatrix}
	=
	\begin{bmatrix}
		\tau_v \\
		\tau_\omega \\
		\tau_\theta
	\end{bmatrix}
\end{equation}
or for short,
\begin{equation}\label{eq:gen_eom_simple}
	Mw+d=\tau
\end{equation}
\Add{where $M$, $w$, $d$ and $\tau$ are a generalized mass matrix, a generalized acceleration, a second order term of a generalized velocity and generalized force respectively. In addition, the subscripts $(\cdot)_v$, $(\cdot)_\omega$ and $(\cdot)_\theta$ indicate translational velocity, rotational velocity and joint speed respectively.}
The corresponding generalized velocities are the translational velocity of the CoM of the entire spacecraft $v_v$, the angular velocity of the main body $v_\omega$, and the joint actuation speed $v_\theta$, which are all expressed in the body frame:
\begin{equation}
    v_v = v_\crm =
    \begin{bmatrix}
      v_{\crm, x}\\ v_{\crm, y}\\ v_{\crm, z}
    \end{bmatrix} \in \R^3 \qquad
    v_\omega = \omega =
    \begin{bmatrix}
      \omega_{x}\\ \omega_{y}\\ \omega_{z}
    \end{bmatrix} \in \R^3 \qquad
    v_\theta = \dot{\theta} =
    \begin{bmatrix}
      \dot{\theta}_1 \\ \vdots \\ \dot{\theta}_m
    \end{bmatrix} \in \R^m\\
\end{equation}
\Add{where $m$ is the number of hinge joints of a transformable spacecraft.}
An explicit description of the components in Eq. (\ref{eq:gen_eom}) is given in Appendix A. The important point is that the terms $M_{v\omega}$, $M_{v\theta}$, $M_{\omega v}$, and $M_{\theta v}$ are all zero in this expression, which means translational and rotational motions are not coupled and independently solved. In this paper, our interest is on attitude motion, and therefore the first row of Eq. \eqref{eq:gen_eom} does not appear in the following discussions. 
In addition, this paper assumes that control input of the system is given as joint acceleration $w_\theta$ and required joint torque $\tau_\theta$ is ideally provided to realize the desired joint acceleration. Therefore, the fundamental equation handled in this paper is the second row of Eq. \eqref{eq:gen_eom}:
\begin{equation}
    \begin{split}
        & M_{\omega\omega}w_\omega + M_{\omega\theta}w_\theta + d_\omega = \tau_\omega \\
        \leftrightarrow\quad 
        & M_{\omega\omega}\dot{\omega} + M_{\omega\theta}\ddot{\theta} + d_\omega = \tau_\omega
    \end{split}
\end{equation}
Moreover, the total angular momentum $h_\crm$ around the centroid of the entire spacecraft can be represented as follows:
\begin{equation}\label{eq:am}
  \begin{split}
		h_\crm &= M_{\omega \omega}v_\omega + M_{\omega \theta}v_\theta
        = M_{\omega \omega}\omega + M_{\omega \theta}\dot{\theta} \\
  \end{split}
\end{equation}
The first term is angular momentum of rotation of an entire spacecraft and the second term is angular momentum of relative rotation of body components.

\subsection{Formulation of solar radiation pressure}\label{sec:srp}
\begin{figure}[htb!]
	\centering\includegraphics[width=4.0in]{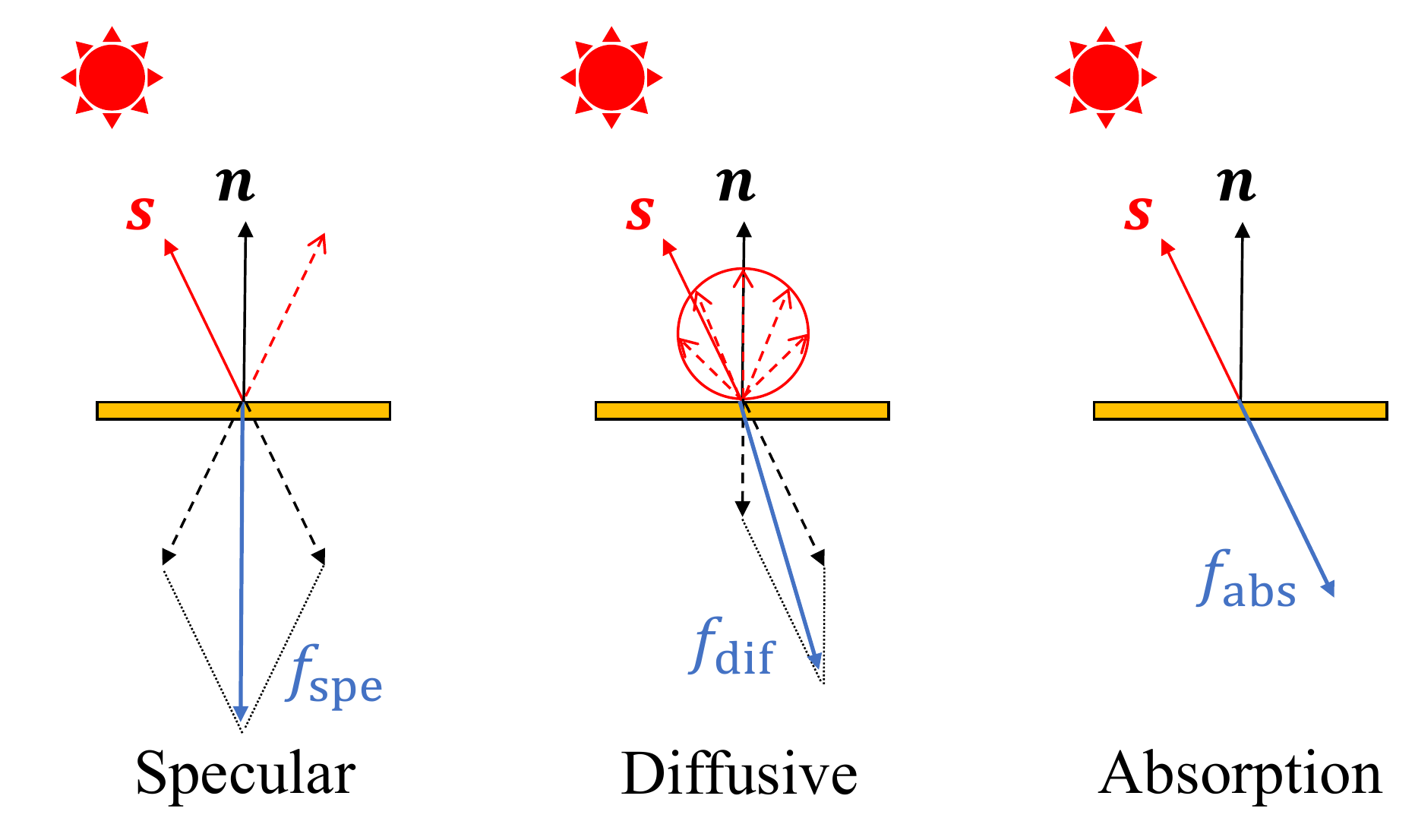}
	\caption{Three factors for SRP force (from left): specular, diffusive, absorptive}
  \label{fig:srp_illust}
\end{figure}

Solar radiation pressure (SRP) is pressure applied on a sunlit surface. Current standard formulation decomposes the SRP into three major factors: specularity, diffusion, and absorption \cite{mcinnes2004solar}. Each factor is illustrated in Fig. \ref{fig:srp_illust}. \par
\Del{In this paper, we set one assumption that all body components of a transformable solar sail are flat and thin. This assumption ensures SRP force is only exerted on the flat surface and the applied force can be determined by a normal vector of the surface. Under this assumption, SRP force exerted on an $i$-th body is expressed as:} 
\Add{In this paper, we set an assumption that the transformable spacecraft is composed only of flat surfaces. Under this assumption, the SRP force exerted on a $j$-th surface of an $i$-th body is characterized by the area of the surface, direction of its (outward-pointing) normal vector and its optical properties as:}
\begin{equation}\label{eq:srp_def}
    \begin{split}
        \Add{F_{ij}} & \Add{= } 
        \begin{cases}
            \Add{\hat{F}_{ij}} & \Add{(n_{ij}^\T s \ge 0)} \\
            \Add{0} & \Add{(n_{ij}^\T s < 0)}
        \end{cases}\\
        \Rep{F_i}{\hat{F}_{ij}}(s,\Rep{n_i}{n_{ij}}) &= - P\Rep{A_i}{A_{ij}} \left(\Rep{n_i}{n_{ij}}^\T s\right)\left\{ (\Rep{C_{\mathrm{abs}i}}{C_{\mathrm{abs}ij}}+\Rep{C_{\mathrm{dif}i}}{C_{\mathrm{dif}ij}})s \right.\\
        &\quad\left.
        + \left(\frac{2}{3}\Rep{C_{\mathrm{dif}i}}{C_{\mathrm{dif}ij}}+2 \left(\Rep{n_i}{n_{ij}}^\T s\right) \Rep{C_{\mathrm{spe}i}}{C_{\mathrm{spe}ij}} \right) \Rep{n_i}{n_{ij}} \right\} \\
        &= - \qty(\Rep{p_{\nrm1i}}{p_{\nrm1ij}} \Rep{n_{1i}}{n_{1ij}} 
        + \Rep{p_{\nrm2i}}{p_{\nrm2ij}}\Rep{n_{2i}}{n_{2ij}}
        + \Rep{p_{\srm i}}{p_{\srm ij}} \Rep{s_i}{s_{ij}} )\Rep{A_i}{A_{ij}} \\
        \text{where} \quad
        \Rep{p_{\nrm1i}}{p_{\nrm1ij}} &= 2P \Rep{C_{\mathrm{spe}i}}{C_{\mathrm{spe}ij}}, \quad
        \Rep{p_{\nrm2i}}{p_{\nrm2ij}} = \frac{2}{3}P \Rep{C_{\mathrm{dif}i}}{C_{\mathrm{dif}ij}}, \quad
        \Rep{p_{\srm i}}{p_{\srm ij}} = P\qty(\Rep{C_{\mathrm{abs}i}}{C_{\mathrm{abs}ij}} + \Rep{C_{\mathrm{dif}i}}{C_{\mathrm{dif}ij}}),   \\ 
        \Rep{n_{1i}}{n_{1ij}} &= \qty(\Rep{n_i}{n_{ij}}^\T s)^2 \Rep{n_i}{n_{ij}}, \quad
        \Rep{n_{2i}}{n_{2ij}} = \qty(\Rep{n_i}{n_{ij}}^\T s) \Rep{n_i}{n_{ij}}, \quad
        \Rep{s_i}{s_{ij}} = \qty(\Rep{n_i}{n_{ij}}^\T s)s \\
    \end{split}
\end{equation}
\Add{As the first equation indicates, SRP force is exerted on the surface where its normal vector is directed to the sun. Note that the shadow detection computing is not performed in this paper due to limited computational resource.}
$\Rep{C_{\mathrm{spe}i}}{C_{\mathrm{spe}ij}}$, $\Rep{C_{\mathrm{dif}i}}{C_{\mathrm{dif}ij}}$, and $\Rep{C_{\mathrm{abs}i}}{C_{\mathrm{abs}ij}}$ are respectively coefficients for the specularity, diffusion, and absorption of \Rep{the $i$-th surface}{the $j$-th surface of the $i$-th body} that holds $\Rep{C_{\mathrm{spe}i}}{C_{\mathrm{spe}ij}}+ \Rep{C_{\mathrm{dif}i}}{C_{\mathrm{dif}ij}}+\Rep{C_{\mathrm{abs}i}}{C_{\mathrm{abs}ij}}=1$. $\Rep{A_i}{A_{ij}}$ is an area of \Rep{the $i$-th flat surface}{the $j$-th surface of the $i$-th body}.
$s\in\R^3$ and $\Rep{n_i}{n_{ij}}\in\R^3$ are components of the vectors towards the sun from the surface and the normal vector of \Rep{the $i$-th surface}{the $j$-th surface of the $i$-th body} respectively (both are unit vectors)\Del{ that holds $n_i^\T s >0$}. $P$ is the solar radiation pressure applied on a certain surface expressed as follows:
\begin{equation}
    P = \frac{S_0}{c}\qty(\frac{1}{\mathrm{AU}})^2
\end{equation}
where $S_0$, $c$, $\mathrm{AU}$ are a solar constant, a speed of light, and a distance between the sun and \Rep{the solar sail}{a spacecraft} expressed in an astronomical unit. Total force and torque applied on a spacecraft are:
\begin{equation}
    \begin{split}
        F &= \sum_i \Add{\sum_j} \Rep{F_i}{F_{ij}} \\
        T &= \sum_i \Add{\sum_j} \Rep{T_i}{T_{ij}} 
        = \sum_i \Add{\sum_j} \Rep{r_{i\crm}^\times}{\qty(r_{\srm_{ij} i}+r_{i\crm})^\times} \Rep{F_i}{F_{ij}} 
        = \sum_i \Add{\sum_j} \qty(\Rep{R_i}{R_{\srm_{ij}}}-R_\crm)^\times \Rep{F_i}{F_{ij}}
    \end{split}
\end{equation}
\Add{Again, the characters appearing in this equation can be interpreted with the nomenclature list at the beginning of this paper. $r_{\srm_{ij} i}$ is the relative position of the center of the $j$-th surface of the $i$-th body with respect to the CoM of the $i$-th body, $r_{i\crm}$ is the relative position of the CoM of the $i$-th body with respect to the CoM of an entire spacecraft. $x^\times$ is a skew symmetric matrix of a component vector $x\in\R^3$.}

\subsection{Jacobian of SRP torque}\label{sec:jac_srp}
\begin{figure}[htb!]
	\centering\includegraphics[width=4.0in]{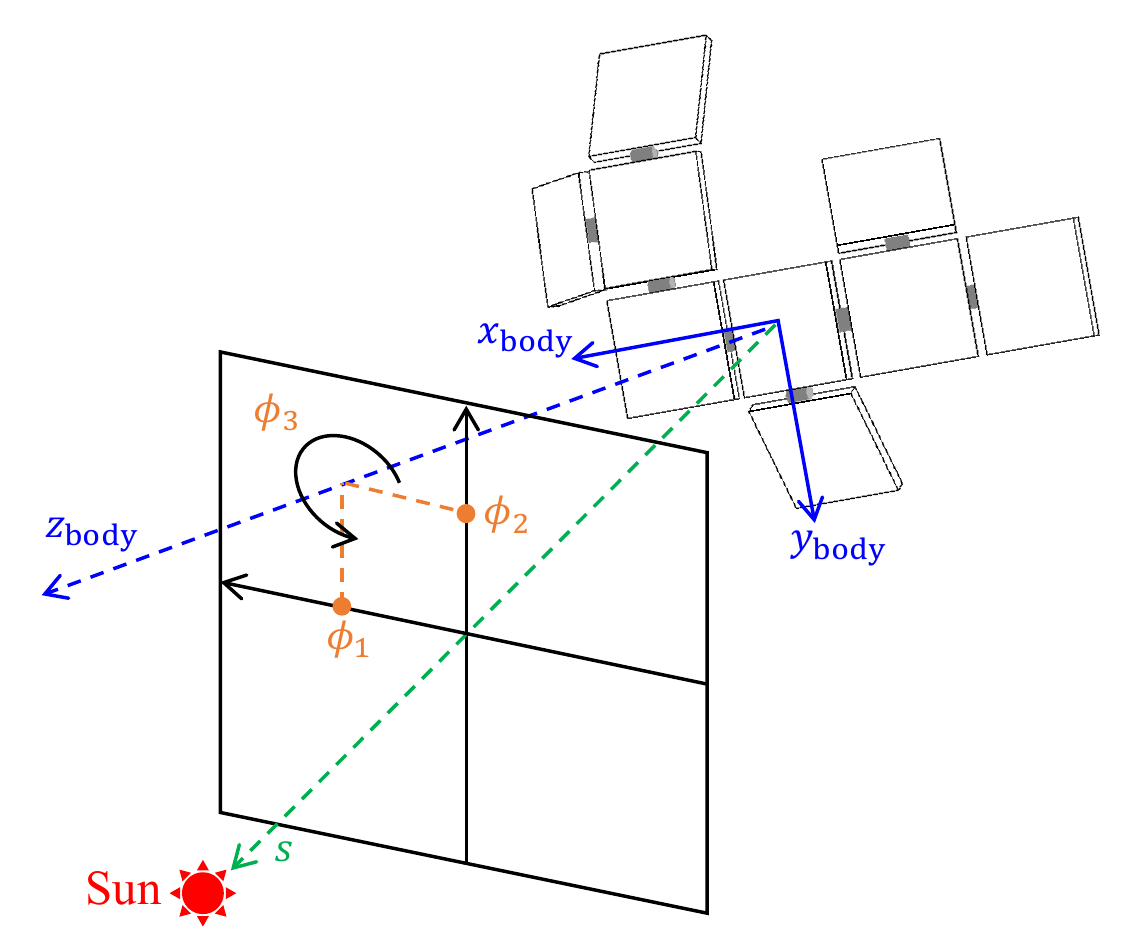}
	\caption{Definitions of coordinate related to attitude motion}
  \label{fig:att_coord}
\end{figure}

Difference of SRP torque is important for analytical discussions and designing stabilization controls. This section is devoted to derivation of the Jacobian. The SRP torque applied on a spacecraft is dependent on attitude and joint angles. Thus, its total derivative is expressed as:
\begin{equation}\label{eq:dt}
    \diff T=\pdv{T}{\phi}\diff\phi + \pdv{T}{\theta}\diff \theta
\end{equation}
where attitude is expressed with Euler angles $\phi=[\phi_1,\ \phi_2,\ \phi_3]$. The following analysis can be applied to any Euler sequence, but 2-1-3 sequence is adopted in this paper. The relationships of the coordinate are shown in Fig. \ref{fig:att_coord}.
Components of sun-direction vector expressed in body fixed frame is:
\begin{equation}\label{eq:s_def}
    s=
    \begin{bmatrix}
        \cos\phi_1\sin\phi_2\sin\phi_3 - \sin\phi_1\cos\phi_3 \\
        \cos\phi_1\sin\phi_2\cos\phi_3 + \sin\phi_1\sin\phi_3 \\
        \cos\phi_1\cos\phi_2
    \end{bmatrix} 
\end{equation}
\Add{Jacobians of $n_{2ij}$ and $s_{ij}$ are discontinuous at $n_{ij}^\T s=0$. To avoid this discontinuity when calculating control input, we set two assumptions. First, we assume all body components have a front face to receive SRP dominantly. This assumption is valid in most cases because a transformable spacecraft should consist of panel-like structures if it is designed to effectively use SRP and to compactly be stowed on a launch vehicle. Second, it is assumed that the front face always satisfies $n_{ij}^\T s>0$ during a simulation. This assumption is also valid in most cases because the proposed control law can be effective when a spacecraft have small and slow attitude oscillation around a sun-facing nominal attitude. These assumptions can prevent the Jacobian from being discontinuous, but rather generate modeling error of SRP. Validity of the control law under this modeling error is confirmed in Section \ref{sec:sim}. Hereafter, the $n_{1ij}$, $n_{2ij}$ and $s_{ij}$ of the front face are denoted as $n_{1i}$, $n_{2i}$ and $s_{i}$. Under these assumptions,} Jacobian of $n_{1i}$, $n_{2i}$, $s_i$ with respect to Euler angles $\phi$ are expressed as follows: 
\begin{equation}\label{eq:dns_dphi}
    \begin{split}
        &\pdv{n_{1i}}{\phi_j} = 2 \qty(n_i^\T s) \qty(n_i^\T \pdv{s}{\phi_j}) n_i \\
        &\pdv{n_{2i}}{\phi_j} = \qty(n_i^\T \pdv{s}{\phi_j}) n_i \\
        &\pdv{s_{i}}{\phi_j} = \qty(n_i^\T \pdv{s}{\phi_j}) s
        + \qty(n_i^\T s) \pdv{s}{\phi_j} \\
        & \pdv{s}{\phi_1} = 
        \begin{bmatrix}
            -\sin\phi_1\sin\phi_2\sin\phi_3 - \cos\phi_1\cos\phi_3 \\
            -\sin\phi_1\sin\phi_2\cos\phi_3 + \cos\phi_1\sin\phi_3 \\
            -\sin\phi_1\cos\phi_2
        \end{bmatrix} \\
        & \pdv{s}{\phi_2} = 
        \begin{bmatrix}
            \cos\phi_1\cos\phi_2\sin\phi_3 \\
            \cos\phi_1\cos\phi_2\cos\phi_3 \\
            -\cos\phi_1\sin\phi_2
        \end{bmatrix} \\
        & \pdv{s}{\phi_3} = 
        \begin{bmatrix}
            \cos\phi_1\sin\phi_2\cos\phi_3 + \sin\phi_1\sin\phi_3 \\
            -\cos\phi_1\sin\phi_2\sin\phi_3 + \sin\phi_1\cos\phi_3\\
            0
        \end{bmatrix}
    \end{split}
\end{equation}
where $\pdv{X}{\phi_j}$ indicates $j$-th column of $\pdv{X}{\phi}$. 
Jacobian of SRP torque with respect to Euler angles is provided as:
\begin{equation}\label{eq:dft_dphi}
    \begin{split}
        &\pdv{T}{\phi_j} = \sum_i \qty(\Rep{R_i}{R_{\srm_i}}-R_\crm)^\times \pdv{F_i}{\phi_j} \\
        \text{where}\quad
        &\pdv{F_i}{\phi_j} = -\qty(p_{\nrm1i}\pdv{n_{1i}}{\phi_j} 
        + p_{\nrm2i}\pdv{n_{2i}}{\phi_j} 
        + p_{\srm i} \pdv{s_{i}}{\phi_j} )A_i \\
    \end{split}
\end{equation}
\Add{where $F_{ij}$, $R_{\srm_{ij}}$ and $A_{ij}$ of the front face are denoted as $F_{i}$, $R_{\srm_{i}}$ and $A_i$.}
On the other hand, Jacobian with respect to joint angles can be computed with the following procedure. Derivative of torque on $i$-th body with respect to angular displacement of $k$-th joint is:
\begin{equation}\label{eq:dt_dthk}
    \pdv{T_i}{\theta_k} = \qty(\frac{\partial \Rep{R_i}{R_{\srm_i}}}{\partial \theta_k}-\pdv{R_\crm}{\theta_k})^\times F_i + \qty(\Rep{R_i}{R_{\srm_i}}-R_\crm)^\times \pdv{F_i}{n_i} \pdv{n_i}{\theta_k}
\end{equation}
The difference of CoM positions, $\pdv{R_\crm}{\theta_k}$ and $\pdv{R_i}{\theta_k}$ are expressed as:
\begin{equation} \label{eq:dr_dthk}
    \begin{split}
        \pdv{R_\crm}{\theta_k} &= \frac{\hat{m}_k}{m}\lambda_k^\times r_{\hat{k}\hrm_k} \\
        \pdv{\Rep{R_i}{R_{\srm_i}}}{\theta_k} &= 
        \begin{cases}
            \lambda_k^\times \Rep{r_{i\hrm_k}}{r_{\srm_i\hrm_k}} & (i\in\hat{k}) \\
            0 & (i\notin \hat{k})
        \end{cases}
    \end{split}
\end{equation}
Derivative of SRP force with respect to difference of $i$-th normal vector is:
\begin{equation}
    \begin{split}
        \pdv{F_i}{n_i} &= -PA_i \qty{  (C_{\mathrm{abs}i}+C_{\mathrm{dif}i})s 
    	+  \left(\frac{2}{3}C_{\mathrm{dif}i}+2 \left(n_i^\T s\right) C_{\mathrm{spe}i}\right) n_i } s^\T  \\
    	&\qquad\qquad\quad
            - PA_i \qty(n_i^\T s) \qty{ 2 C_{\mathrm{spe}i} n_i s^\T + \left(\frac{2}{3}C_{\mathrm{dif}i}+2 \left(n_i^\T s\right) C_{\mathrm{spe}i}\right)U} \\
    	&= -PA_i \left\{ (C_{\mathrm{abs}i}+C_{\mathrm{dif}i})ss^\T
    	+ \left(\frac{2}{3}C_{\mathrm{dif}i}+4 \left(n_i^\T s\right) C_{\mathrm{spe}i}\right) n_i s^\T \right.\\
    	&\qquad\qquad\quad \left.
            + \qty(n_i^\T s) \qty(\frac{2}{3}C_{\mathrm{dif}i}+2 \left(n_i^\T s\right) C_{\mathrm{spe}i}) U \right\}
    \end{split}
\end{equation}
The last term, diffrence of normal vector of $i$-th body with respect to angular displacement of $k$-th joint is:
\begin{equation}\label{eq:dn_dthk}
    \begin{split}
        \pdv{n_i}{\theta_k} &= 
        \begin{cases}
            \lambda_k^\times n_i & (i\in\hat{k}) \\
            0 & (i\notin \hat{k})
        \end{cases}
    \end{split}
\end{equation}
\Del{Equation \eqref{eq:dt_dthk} can be computed by substituting Eqs. \eqref{eq:dr_dthk}--\eqref{eq:dn_dthk}. Finally, total derivative Eq. \eqref{eq:dt} is obtained from the Eqs. \eqref{eq:dns_dphi} and \eqref{eq:dt_dthk}.}
\Add{The derivative of torque on $i$-th body with respect to $k$-th joint actuation can be computed by substituting Eqs. \eqref{eq:dr_dthk}--\eqref{eq:dn_dthk} into Eq. \eqref{eq:dt_dthk}. Finally, the total derivative of the SRP torque on an entire spacecraft is obtained by substituting Eqs. \eqref{eq:dft_dphi} and \eqref{eq:dt_dthk} into Eq. \eqref{eq:dt}.}

\section{Proposed methods} \label{sec:methods}

\subsection{SRP-based joint angle optimization} \label{sec:jopt}
This section describes an optimization problem to obtain optimal attitude and joint angles that satisfies target SRP force and torque while achieving stable attitude motion.
The optimization problem is defined as follows:
\begin{equation}
    \begin{aligned}
    & \underset{\phi,\theta}{\text{minimize}}
    & & f(\phi,\theta) \\
    & \text{subject to}
    & & c(\phi,\theta)\le0,\quad c_\mathrm{eq}(\phi,\theta,F_\targ,T_\targ)=0,\quad \theta_\mathrm{lb}\le\theta\le\theta_\mathrm{ub}
    \end{aligned}
\end{equation}
where $\theta_\mathrm{lb},\ \theta_\mathrm{ub}\in\R^m$ are lower and upper bound of joint angles.
Each function in the above optimization problem is set as follows:
\begin{equation}\label{eq:optim}
    \begin{split}
        f &= -\left\| \Im\qty( \sqrt{\Lambda} ) \right\|_\infty \\
        c &=  \max\qty(\Re\qty( \sqrt{\Lambda} )) \\
        c_\mathrm{eq} &= 
        \begin{bmatrix}
            F_\targ-F \\
            T_\targ-T
        \end{bmatrix}
    \end{split}
\end{equation}
where $\|\cdot\|_\infty$ is an L-infinity norm of vector components, $\Re(\cdot)$ and $\Im(\cdot)$ take real and imaginary part of each component respectively, and $\max(\cdot)$ takes a maximum value of all of the components. $\Lambda$ is a $3\times 1$ matrix that contains all three eigenvalues of the following matrix $A_\phi\in\R^{3\times3}$:
\begin{equation}
    A_\phi = C_\phi I_\crm^{-1} \pdv{T}{\phi}
\end{equation}
where $C_\phi$ is a matrix that converts angular velocity into Euler angle velocity, hence $\dot{\phi}=C_\phi\omega$. 
The objective function $f$ evaluates frequency of attitude oscillation. The time scale of this frequency determines time scale of the damping control described later in Section \ref{sec:damp}, and therefore it is minimized to attain rapid momentum damping. The nonlinear inequality constraint $c$ determines divergence property of the attitude motion, where $c\le0$ means the attitude motion does not diverge. Actually, it is shown by Chujo that $c$ does not become negative under SRP torque field \cite{chujo2022propellant}. However, we rather set $c\le0$ because the convergence performance of the solver is improved by this setting. $c_\mathrm{eq}$ defines constraints for SRP force and torque applied on a spacecraft. In total, a successful optimal solution provides optimal attitude and joint angles that satisfy arbitrary target SRP force and torque while achieving (neutrally) stable attitude motion. We solved the optimization problem defined in Eq. \eqref{eq:optim} by using \texttt{fmincon}, the constrained optimization solver in MATLAB. The algorithm of the solver is set to be the interior point method \cite{byrd1999interior}. \par 
The target $F_\targ$ and $T_\targ$ can be arbitrary values as far as their magnitudes are within the range of available SRP force and torque. However, $T_\targ=[0,\ 0,\ 0]^\T$ is practically used in most cases because it generates (neutrally) stable equilibrium attitude satisfying arbitrary target SRP force, which is useful for orbit control. Therefore, we assume $T_\targ=[0,\ 0,\ 0]^\T$ in the following sections. \par 
In this paper, initial guess of $\phi$ and $\theta$ are provided in the following procedure. First, the initial guess of joint configuration, $\theta_0$, is provided as the body configuration exhibits an "umbrella-like" shape against the sun; joints are folded such that the CoM of the body component moves opposite to the $\vect{s}$ direction. It has been pointed out that this umbrella-like body configuration improves stability of attitude oscillation under SRP torque field \cite{kubo2019propellant}. 
Next, the initial guess of Euler angles, $\phi_0$ is provided as a normal vector of the body-$0$ points opposite to the direction of target SRP force. This initial guess is expected to make a direction of total SRP force be close to that of the target SRP force. Roll angle around the normal vector is not uniquely determined with this initial guess, and therefore, we adopts trial-and-error search; the roll angle from $0$ to $2\pi$ is tested in a certain interval until the solver returns a successful (local) optimal solution.

\subsection{Momentum damping control with joint actuation} \label{sec:damp}
As Chujo pointed out \cite{chujo2022propellant}, the optimal solution obtained in Section \ref{sec:jopt} cannot stabilize rotation around the sun-pointing vector $\vect{s}$ because this rotation does not affect SRP torque expressed in body fixed frame. Therefore, minute residual angular velocity or angular acceleration around the vector $\vect{s}$ are never decelerated and generates attitude drift from the desired attitude. Chujo proposed damping control using reaction wheels, but this paper proposes damping control only with joint angle actuation. To construct this control law, the attitude equation of motion is extended into attitude-joint coupled equation as it also includes the counter rotation of attitude during joint actuation. \par
In the following formulations, the superscript $(\cdot)^\mathrm{I}$ such as $T^\Irm$ and $h_\crm^\Irm$ indicates vector components expressed in inertial frame. 
Equation of motion that includes attitude reaction during joint actuation is expressed as follows using Eq. \eqref{eq:am}:
\begin{equation}
    \begin{split}
        T^\Irm &= \dv{h_\crm^\Irm}{t} \\
        \leftrightarrow\quad
        C^\T T &= \dv{t} \qty(C^\T h_\crm) 
        = \dv{t} \qty(C^\T M_{\omega\omega}\omega 
        + C^\T M_{\omega\theta}\dot{\theta}) \\
        & = C^\T \qty{
        \omega^\times M_{\omega\omega}\omega
        + \qty(\pdv{M_{\omega\omega}}{\theta}\dot{\theta})\omega 
        + M_{\omega\omega}\dot{\omega}
        + \omega^\times M_{\omega\theta}\dot{\theta}
        +\qty(\pdv{M_{\omega\theta}}{\theta}\dot{\theta})\dot{\theta}
        + M_{\omega\theta}\ddot{\theta} } \\
        &\simeq C^\T \qty(
        M_{\omega\omega}\dot{\omega}
        + M_{\omega\theta}\ddot{\theta} ) \\
        \leftrightarrow\quad
        T &= M_{\omega\omega}\dot{\omega}
        + M_{\omega\theta}\ddot{\theta}
        \Rep{=}{\simeq} M_{\omega\omega}C_\phi^{-1}\ddot{\phi}
        + M_{\omega\theta}\ddot{\theta}
    \end{split}
\end{equation}
where $C$ is a direction cosine matrix of body frame with respect to inertial frame. Here we assumed $|\omega|, \qty|\dot{\theta}|<<1$ and second order terms of them are ignored.
Now, this equation is linearized around equilibrium state to construct the control law. The equilibrium states $\tilde{T}$, $\tilde{\phi}$, and $\tilde{\theta}$ and difference from the equilibrium states $\delta T$, $\delta\phi$, and $\delta \theta$ satisfy the following equations:
\begin{equation}
    T=\tilde{T}+\delta T=\delta T,\quad \phi=\tilde{\phi}+\delta \phi,\quad  \theta=\tilde{\theta}+\delta \theta
\end{equation}
$\delta T$ can be expressed as follows according to Eq. \eqref{eq:dt}:
\begin{equation}
    \delta T(\delta \phi, \delta \theta) 
    = \pdv{T}{\phi} \qty(\tilde{\phi}, \tilde{\theta}) \delta \phi 
    + \pdv{T}{\theta} \qty(\tilde{\phi}, \tilde{\theta}) \delta \theta 
\end{equation}
Thus, the equation of motion can be linearized around equilibrium state as follows:
\begin{equation}
    \begin{split}
        & \delta T = \pdv{T}{\phi}\delta \phi + \pdv{T}{\theta}\delta \theta 
        = M_{\omega\omega}C_\phi^{-1}\delta\ddot{\phi}
        + M_{\omega\theta}\delta\ddot{\theta} \\
        \leftrightarrow \quad 
        & \delta \ddot{\phi} = C_\phi M_{\omega\omega}^{-1} \pdv{T}{\phi}\delta \phi 
        + C_\phi M_{\omega\omega}^{-1} \pdv{T}{\theta}\delta \theta 
        - C_\phi M_{\omega\omega}^{-1} M_{\omega\theta}\delta\ddot{\theta}
    \end{split}
\end{equation}
Finally, a linearized equation of motion of the attitude-joint coupled motion is provided as follows:
\begin{equation}
    \begin{split}
        \dv{t}
        \begin{bmatrix}
            \delta \phi \\ \delta \theta \\ \delta \dot{\phi} \\ \delta \dot{\theta}
        \end{bmatrix}
        &=
        \begin{bmatrix}
            O_{3\times 3} & O_{3\times m} & U_{3\times 3} & O_{3\times m} \\
            O_{m\times 3} & O_{m\times m} & O_{m\times 3} & U_{m\times m} \\
            C_\phi M_{\omega\omega}^{-1} \pdv{T}{\phi} & C_\phi M_{\omega\omega}^{-1} \pdv{T}{\theta} & O_{3\times 3} & O_{3\times m} \\
            O_{m\times 3} & O_{m\times m} & O_{m\times 3} & O_{m\times m}
        \end{bmatrix}
        \begin{bmatrix}
            \delta \phi \\ \delta \theta \\ \delta \dot{\phi} \\ \delta \dot{\theta}
        \end{bmatrix}
        + 
        \begin{bmatrix}
            O_{3\times m} \\ O_{m\times m} \\ -C_\phi M_{\omega\omega}^{-1} M_{\omega\theta} \\ U_{m\times m}
        \end{bmatrix} \delta\ddot{\theta} \\
        \leftrightarrow\quad 
        \dv{x}{t} &= Ax+Bu 
    \end{split}
\end{equation}
where 
\begin{equation}
    \begin{split}
        A=
        \begin{bmatrix}
            O_{3\times 3} & O_{3\times m} & U_{3\times 3} & O_{3\times m} \\
            O_{m\times 3} & O_{m\times m} & O_{m\times 3} & U_{m\times m} \\
            C_\phi M_{\omega\omega}^{-1} \pdv{T}{\phi} & C_\phi M_{\omega\omega}^{-1} \pdv{T}{\theta} & O_{3\times 3} & O_{3\times m} \\
            O_{m\times 3} & O_{m\times m} & O_{m\times 3} & O_{m\times m}
        \end{bmatrix},\quad
        B=
        \begin{bmatrix}
            O_{3\times m} \\ O_{m\times m} \\ -C_\phi M_{\omega\omega}^{-1} M_{\omega\theta} \\ U_{m\times m}
        \end{bmatrix}
    \end{split}
\end{equation}
\Add{$U$ and $O$ are identity and zero matrices respectively, and the subscript $()_{m\times n}$ denotes that the dimension of the matrix is $m\times n$.}
\Add{In this paper, the control input $u$ is set to be angular acceleration of joints, although the actual motor does not necessarily control angular acceleration.}
To stabilize this system, we adopt the linear-quadratic regulator (LQR). The LQR feedback was proposed by Kalman \cite{kalman1960contributions}, in which an objective function $f$ is provided as quadratics of state and input for linearized system as follows:
\begin{equation}
    \begin{split}
        f = \int_0^\infty \qty( x^\T Q x + u^\T R u )\diff t
    \end{split}
\end{equation}
Then, the optimal feedback control law is provided as follows:
\begin{equation}
    \begin{split}
        u = -R^{-1}B^\T X x
    \end{split}
\end{equation}
where $X$ is a solution of Riccati equation $XA+A^\T X - XBR^{-1}B^\T X+Q=0$. The weight matrices $Q$ and $R$ are provided as follows:
\begin{equation}
    \begin{split}
        Q&=
        \begin{bmatrix}
            Q_{11}&& O& \\ 
            &Q_{22}&& \\
            &&Q_{33}& \\
            &O&&Q_{44}
        \end{bmatrix},\quad
        R=
        \begin{bmatrix}
            |\delta \ddot{\theta}_1|^{-2}& & O\\ 
            &\ddots& \\
            O && |\delta \ddot{\theta}_m|^{-2}
        \end{bmatrix} \\
        Q_{11} &= 
        \begin{bmatrix}
            |\delta \phi_1|^{-2}& & O\\ 
            &|\delta \phi_2|^{-2}& \\
            O && |\delta \phi_3|^{-2}
        \end{bmatrix},\quad
        Q_{22} = 
        \begin{bmatrix}
            |\delta \theta_1|^{-2}& & O\\ 
            &\ddots& \\
            O && |\delta \theta_m|^{-2}
        \end{bmatrix}\\
        Q_{33} &= 
        \begin{bmatrix}
            |\delta \dot{\phi}_1|^{-2}& & O\\ 
            &|\delta \dot{\phi}_2|^{-2}& \\
            O && |\delta \dot{\phi}_3|^{-2}
        \end{bmatrix},\quad
        Q_{44} = 
        \begin{bmatrix}
            |\delta \dot{\theta}_1|^{-2}& & O\\ 
            &\ddots& \\
            O && |\delta \dot{\theta}_m|^{-2}
        \end{bmatrix}\\
    \end{split}
\end{equation}
where the order of magnitude of each state component is provided as follows: 
\begin{equation}
    \begin{split}
        &|\delta \phi_i| \simeq \frac{\pi}{180}, \quad
        |\delta \theta_j| \simeq \frac{\pi}{180}
        \quad \qty(i=1,2,3,\ j=1,\cdots,m)\\
        &|\delta \dot{\phi}_i| \simeq \omega_\nrm |\delta \phi_i|, \quad
        |\delta \dot{\theta}_j| \simeq \omega_\nrm |\delta \theta_j|, \quad
        |\delta \ddot{\theta}_j| \simeq \omega_\nrm^2 |\delta \theta_j|
    \end{split}
\end{equation}
The natural angular frequency $\omega_\nrm$ can be estimated by frequency of uncontrolled attitude motion under SRP, hence $\omega_\nrm=\left\| \Im\qty( \sqrt{\Lambda} ) \right\|_\infty$.\par
\Add{LQR feedback control is a typical example to stabilize the system, but it is not the unique solution. Application of other effective control laws is in the scope of future works.}

\section{Numerical simulations} \label{sec:sim}
\begin{figure}[hbt!]
	\centering\includegraphics[width=4.0in]{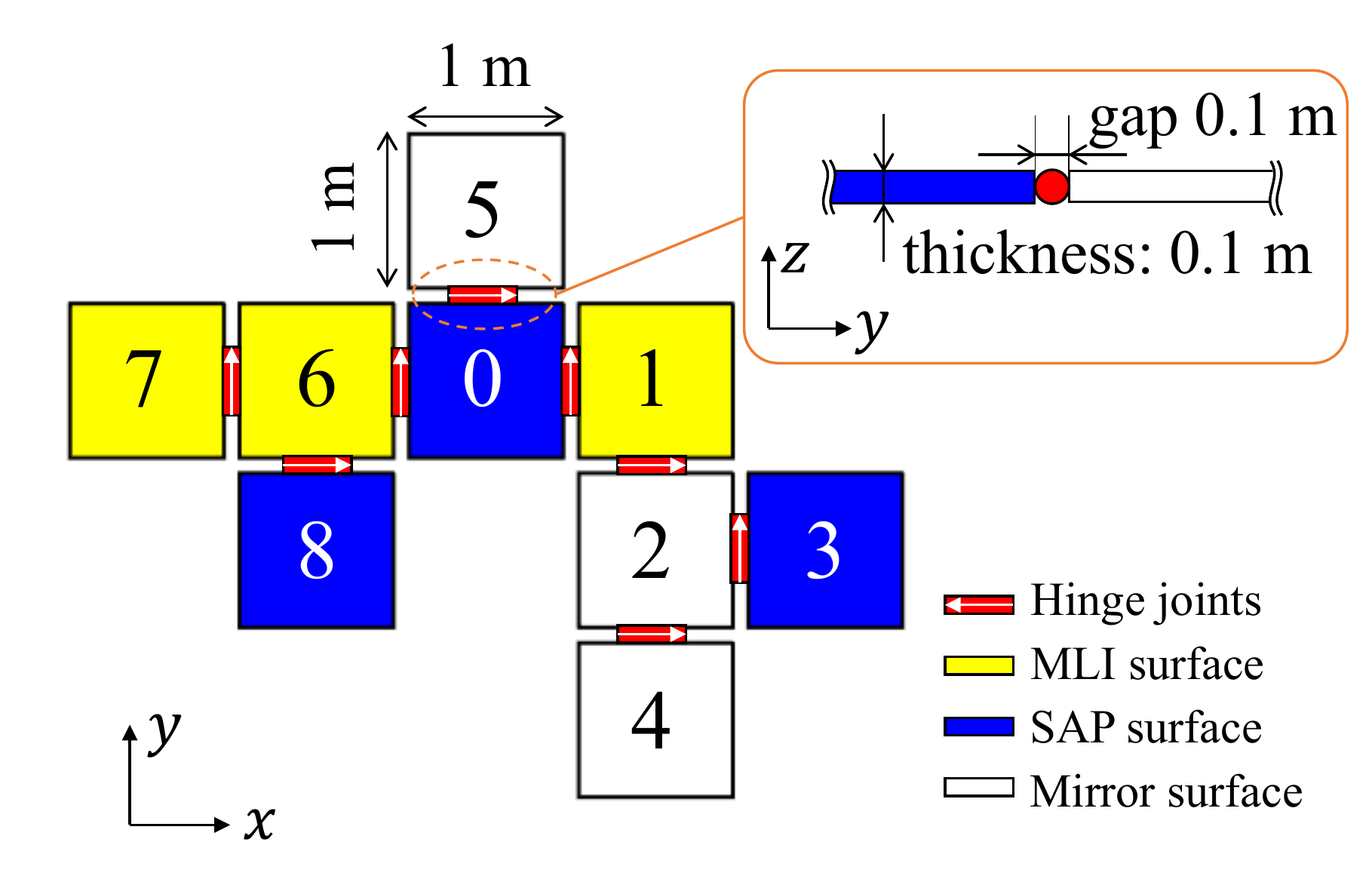}
	\caption{Model of a transformable solar sail used in the simulation}
	\label{fig:scmodel}
\end{figure}

This section provides a numerical simulation to validate the proposed joint optimization and feedback control law to stabilize the attitude motion.
As described in Section \ref{sec:methods}, our proposed methods are \Rep{applicable to an arbitrary transformable solar sail as far as it consists of flat and thin body components}{effective for a transformable spacecraft that has front faces that can dominantly receive SRP on each body}. A mathematical model of a transformable \Rep{solar sail}{spacecraft} used in this paper is shown in Fig. \ref{fig:scmodel}. 
This model consists of nine square flat panels connected with actuatable hinge joints. All panels has identical dimensions; $1\mathrm{m}\times1\mathrm{m}\times0.1\mathrm{m}$ size and $10$ kg mass with homogeneous density. The number of actuatable joints are determined to ensure sufficient control degree of freedom to obtain successful solutions. The configuration shown in this figure exhibits the joint angles are all zero, and local coordinates coincide with the body-fixed coordinate attached to the body $0$ that is shown at the left-bottom part of this figure. The positive direction of joint axis is shown in a white arrow on the hinge joint in Fig. \ref{fig:scmodel}. The gaps between panels are same as thickness of a panel ($0.1$ m), and the axis of joints are located at the middle of the gap. \Del{In the following simulations, we assume SRP is exerted only on the $+z$ surface of each body components.}
Moreover, the panels and its optical properties are randomly distributed to demonstrate that our proposed method is valid for an arbitrary transformable \Rep{solar sail}{spacecraft}. \Add{Three surface materials, multi-layer insulation (MLI), solar array panel (SAP) and ideally reflective mirror are prepared as representative surface materials. Surface materials of bodies are indicated by colors in Fig. \ref{fig:scmodel} where the color labels are described at the right-bottom corner.} The values of optical properties are listed in Table \ref{tab:op}.
\begin{table}[htb]
    \caption{List of optical properties}
    \centering
    \begin{tabular}{cccc}
        \toprule
        Material & $C_{\mathrm{spe}}$ & $C_{\mathrm{dif}}$ & $C_{\mathrm{abs}}$ \\ \hline
        MLI & 0.375 & 0.255 & 0.370 \\
        \Rep{Solar array}{SAP}  & 0.086 & 0.060 & 0.854 \\
        Mirror & 1.0 & 0.0 & 0.0 \\ 
        \bottomrule
    \end{tabular}
    \label{tab:op}
\end{table}
\Add{As is explained in Section \ref{sec:srp}, SRP is exerted on all flat faces directed toward the sun. The control law is constructed so that only Jacobians of front faces are taken into account. The following simulation confirms the control law is valid for this panel-based transformable spacecraft.}

\subsection{SRP-based joint angle optimization}\label{sec:sim_jopt}
\begin{figure}[hbt!]
	\centering\includegraphics[width=4.0in]{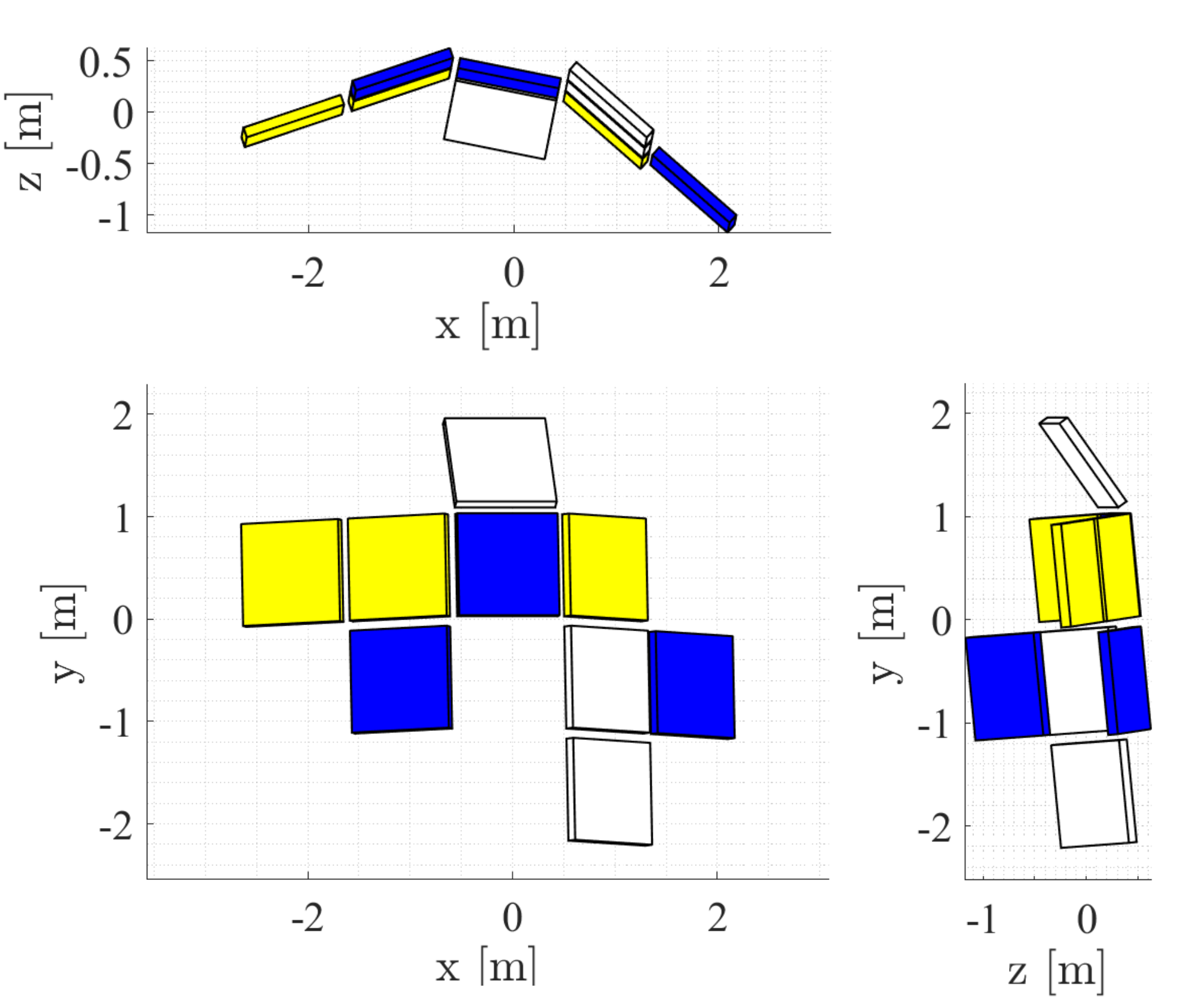}
	\caption{Initial guess of SRP-based joint angle optimization}
	\label{fig:j_opt_initial}
\end{figure}

\begin{figure}[hbt!]
	\centering\includegraphics[width=4.0in]{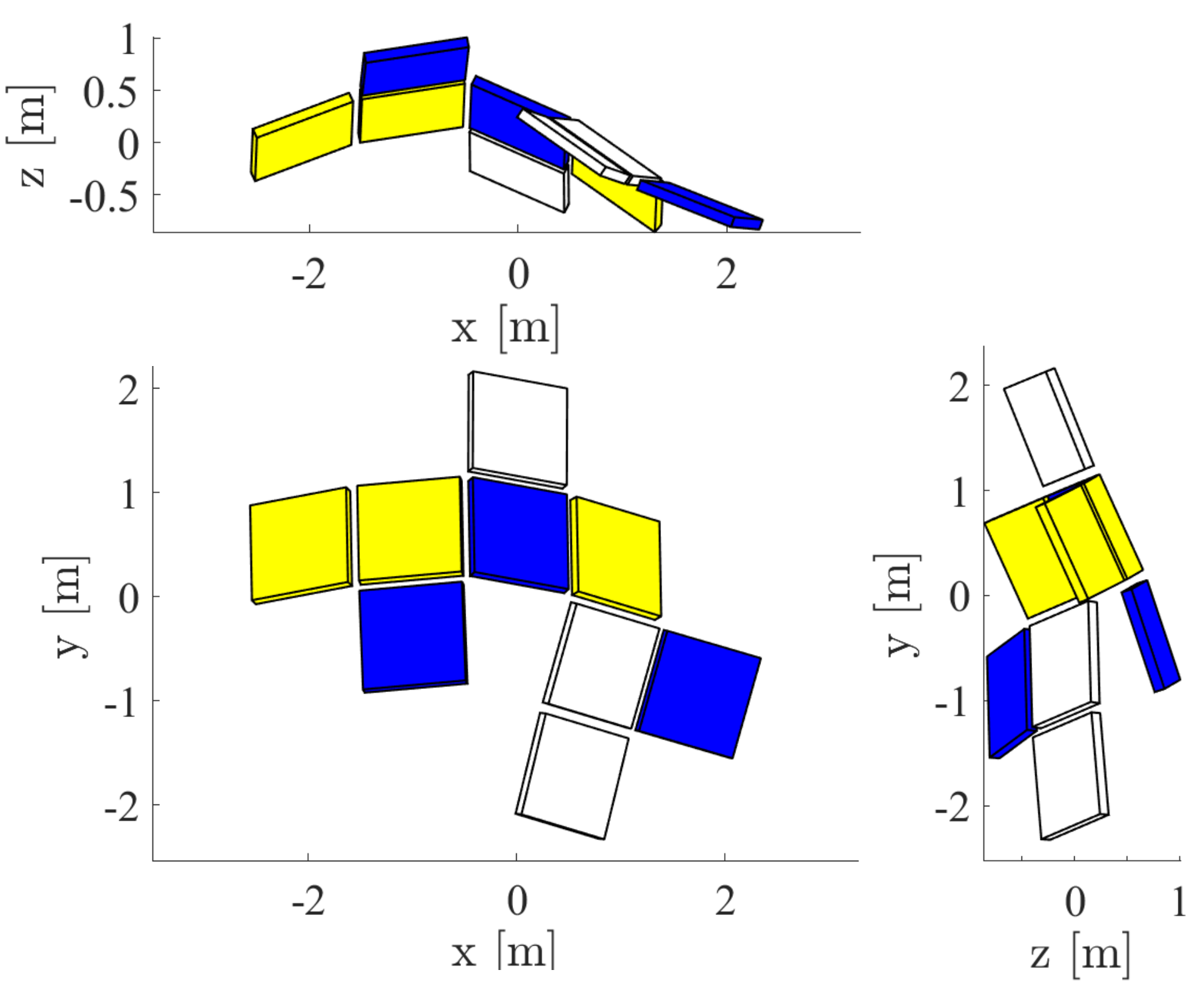}
	\caption{Result of SRP-based joint angle optimization}
	\label{fig:j_opt_result}
\end{figure}

First, the SRP-based joint angle optimization is carried out.
Target SRP force and torque are set as $F_\targ^\Irm = 10^{-4}\times [-0.0868,\ -0.0434,\ -0.4340]^\T$ (N) and $T_\targ=[0,\ 0,\ 0]^\T$ (N$\cdot$m) respectively. Notice that components of the target force is expressed in inertial frame.
Distance from the sun is set as 1.01 AU, which corresponds to the distance between the sun and the Sun-Earth 2nd Lagrange point. 
As mentioned in Section \ref{sec:jac_srp}, we use 2-1-3 Euler angles for numerical integration. 
Initial guess of attitude $\phi$ and joint angles $\theta$ are respectively provided as $\phi_0=[11.31,\ -5.60,\ 0.00]^\T$ (deg) and $\theta_0=[30,\ 0,\ 0,\ 0,\ -30,\ -30,\ 0,\ 0]^\T$ (deg). 
The initial configuration is illustrated in Fig. \ref{fig:j_opt_initial}. In this condition, the optimal attitude and joint configuration obtained by the solver are $\phi=[\Rep{26.79}{27.72},\ \Rep{-26.08}{-22.38},\ \Rep{-14.37}{-10.23}]^\T$ (deg) and $\theta=[\Rep{8.30}{11.87},\ \Rep{26.19}{28.21},\ \Rep{19.45}{-13.14},\ \Rep{8.19}{-3.86},\ \Rep{12.06}{2.79},\ \Rep{-42.86}{-35.39},\ \Rep{-2.72}{-12.89},\ \Rep{26.74}{6.21}]^\T$ (deg) respectively. The result is illustrated in Fig. \ref{fig:j_opt_result}. \par
\subsection{Joint-driven attitude stabilization}
\begin{figure}[hbt!]
	\centering\includegraphics[width=6.0in]{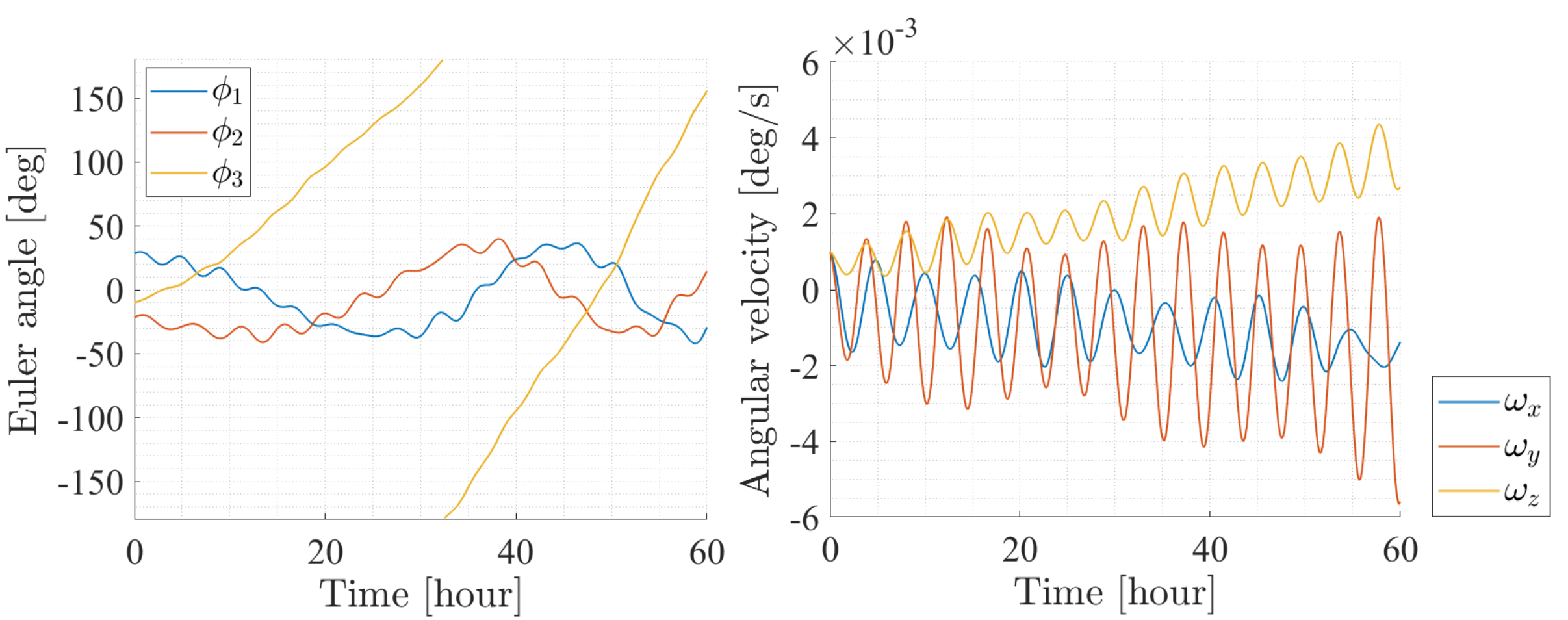}
	\caption{Time history of Euler angles and angular velocity without momentum damping control}
	\label{fig:euler_omg_nocont}
\end{figure}

\begin{figure}[hbt!]
	\centering\includegraphics[width=6.0in]{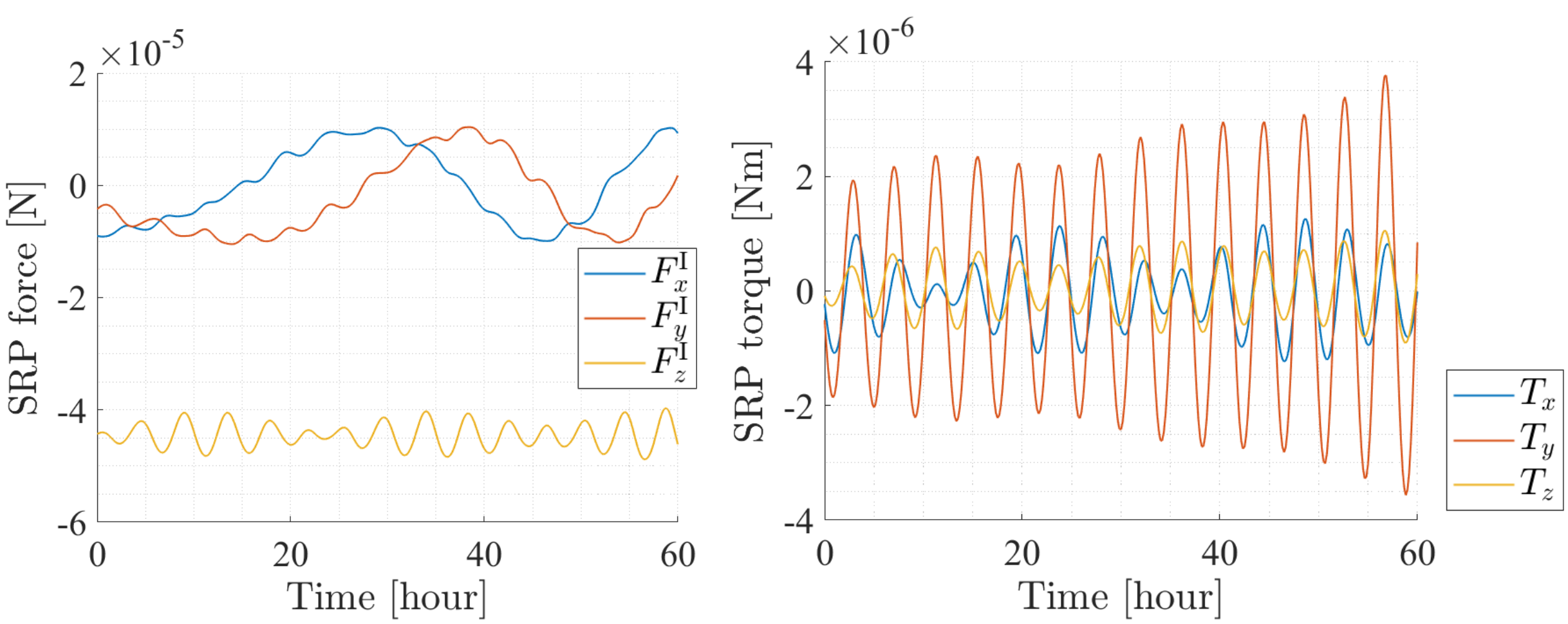}
	\caption{Time history of SRP force and Torque without momentum damping control}
	\label{fig:srp_ft_nocont}
\end{figure}

Now, the stability of the obtained equilibrium state is examined with a simulation. First, the stability without feedback control is examined.
Initial errors of attitude and angular velocity from the equilibrium state are set as $\delta\phi_0=[0.819,\ 0.567,\ 0.088]^\T$ (deg) and $\omega_0=[10^{-3}\ 10^{-3}\ 10^{-3}]^\T$ (deg/s). $\delta\phi_0$ was randomly sampled as the magnitude of the error becomes 1 degree. The numerical integration is carried out with \texttt{ode45} solver in MATLAB with absolute and relative tolerance $10^{-5}$ and $10^{-6}$ respectively. Figure \ref{fig:euler_omg_nocont} and \ref{fig:srp_ft_nocont} show the time histories of Euler angles, angular velocities, and SRP force and torque.
Without momentum damping control, the attitude oscillation around the equilibrium point is never suppressed and gradually drifted from the equilibrium attitude. 
Thus, the result shows the necessity of momentum damping control to maintain the optimal configuration.\par
\begin{figure}[hbt!]
	\centering\includegraphics[width=6.0in]{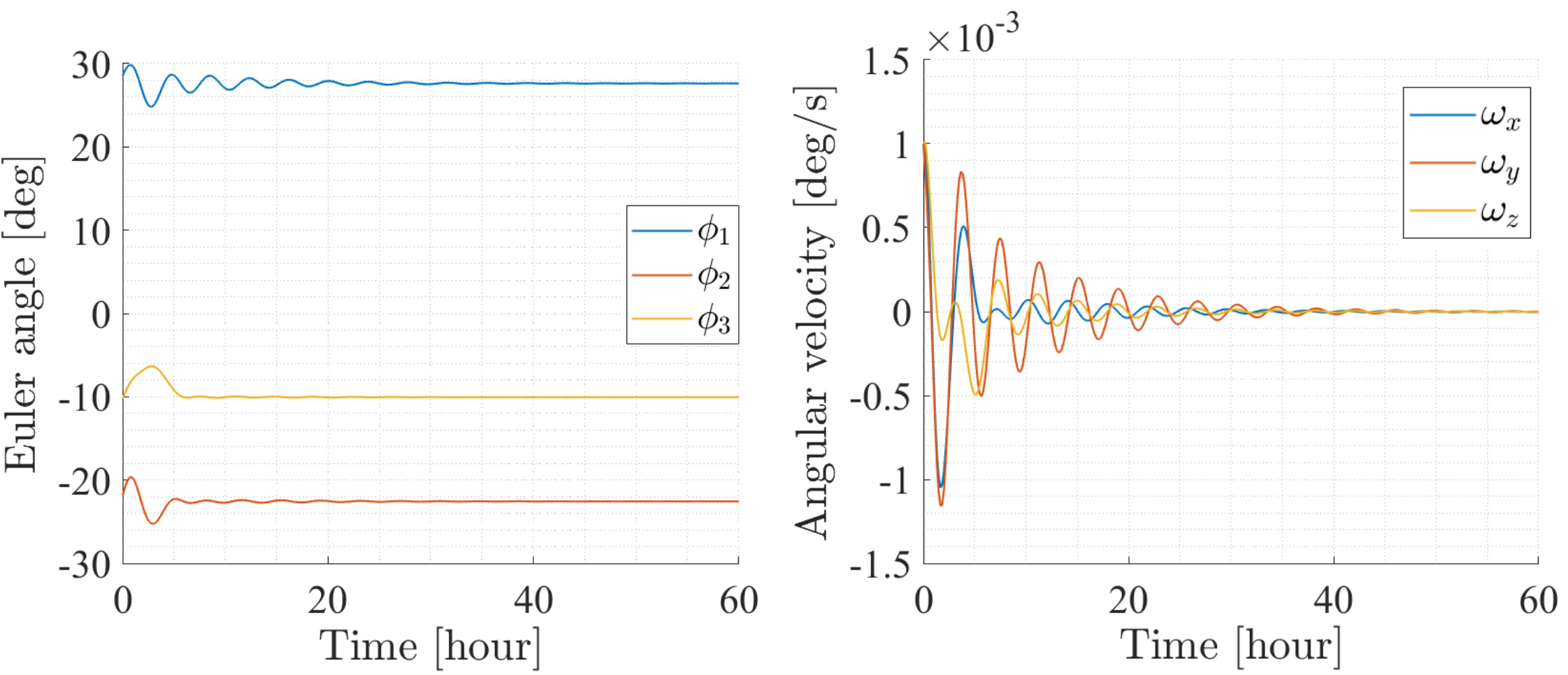}
	\caption{Time history of Euler angles and angular velocity}
	\label{fig:euler_omg}
\end{figure}
\begin{figure}[hbt!]
	\centering\includegraphics[width=6.0in]{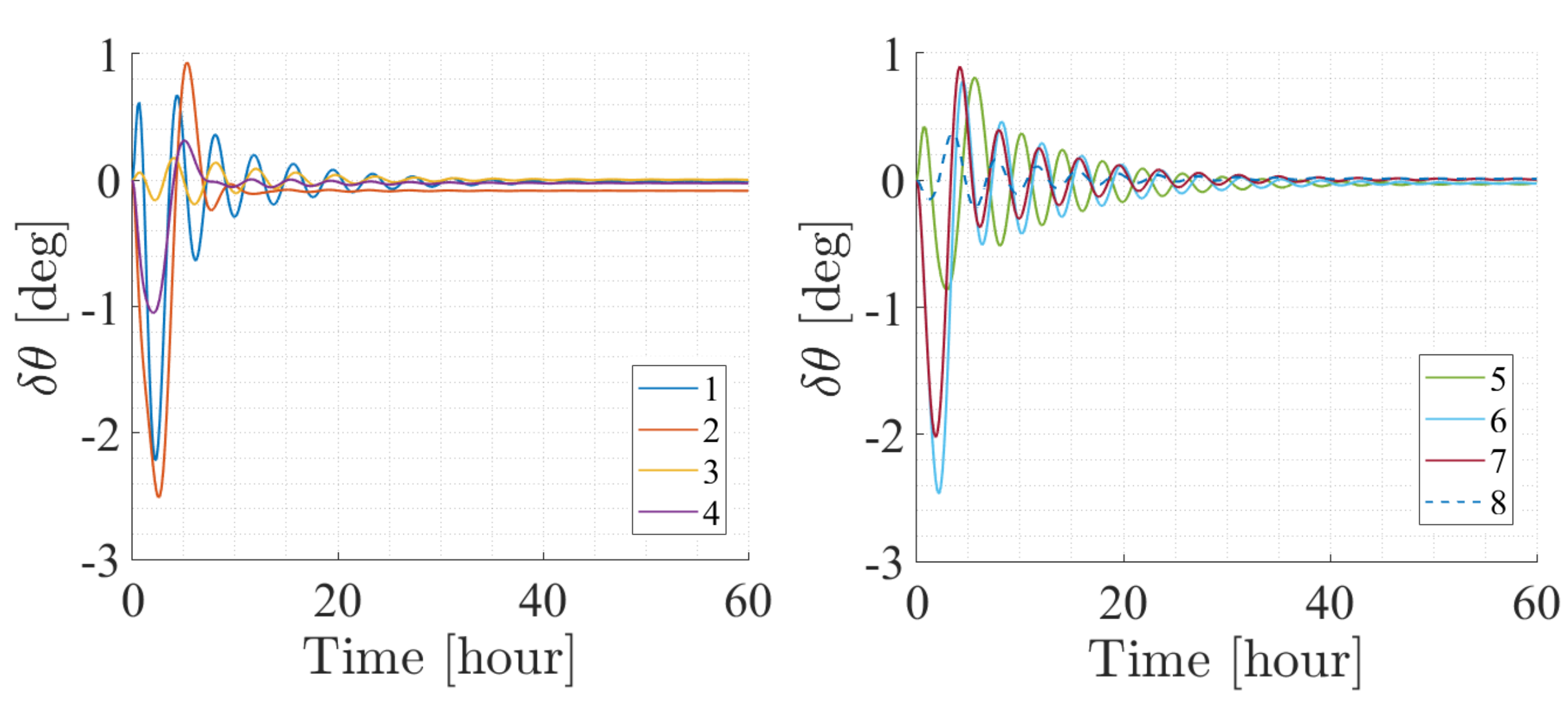}
	\caption{Time history of joint angle difference from the equilibrium configuration}
	\label{fig:j_angles}
\end{figure}
\begin{figure}[hbt!]
	\centering\includegraphics[width=6.0in]{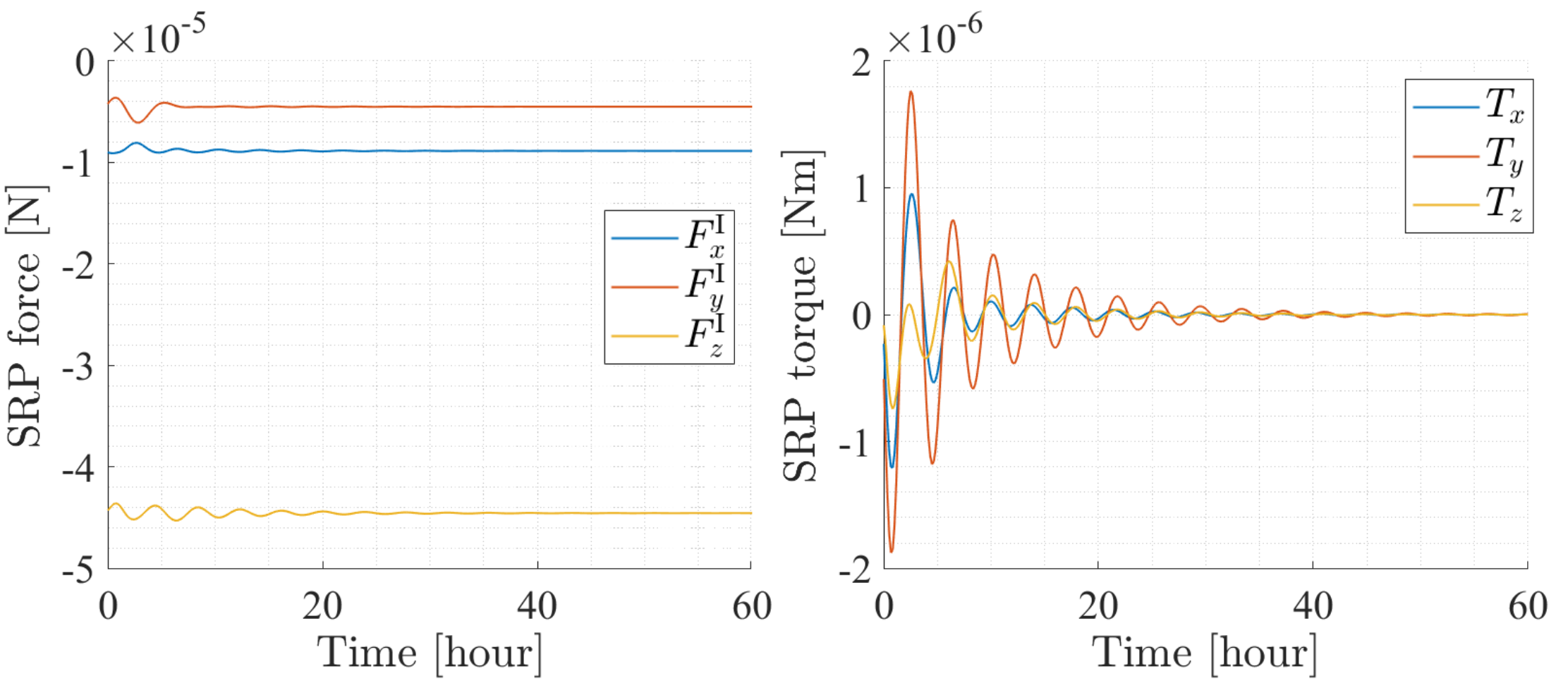}
	\caption{Time history of SRP force and Torque}
	\label{fig:srp_ft}
\end{figure}

Next, momentum damping control is carried out for the system to stabilize the attitude motion. The same initial conditions, $\delta\phi_0=[0.819,\ 0.567,\ 0.088]^\T$ (deg) and $\omega_0=[10^{-3}\ 10^{-3}\ 10^{-3}]^\T$ (deg/s) are also used for this simulation. The natural frequency of the attitude oscillation becomes $\omega_\nrm=\Rep{4.53}{4.23}\times10^{-4}$ (1/s) in this case, and this value is reflected to the weight matrices $Q$, $R$ of the LQR feedback.
Figure \ref{fig:euler_omg}--\ref{fig:srp_ft} show the result of the simulation with the momentum damping control. Figure \ref{fig:j_angles} is expressed as difference from the equilibrium joint angles.
All the figures show that the initial error of the attitude and angular velocity are suppressed and the momentum damping is successfully completed while final values of attitude and joint configuration are converged to the target state obtained in Section \ref{sec:sim_jopt}. As a result, the transformable spacecraft can stably acquire the desired SRP acceleration without consuming any expendable propellant.

\section{Conclusions} \label{sec:concl}
The present work contributed to enhancing orbit control capability of transformable spacecrafts. Though the transformable spacecrafts have high redundancy in control degree of freedom, \Del{but} its large number of control inputs imposed difficulties in control in previous studies. This paper solved this problem with two novel proposed techniques.
First, we proposed the optimization method to obtain attitude and joint configuration to satisfy arbitrary SRP force and torque while preventing divergence of attitude. Second, we proposed the momentum damping control formulated under attitude-joint coupled dynamics. What distinguishes these methods from other previous methods is that they are generally applicable to any transformable \Rep{solar sail}{spacecraft} \Rep{that has flat and thin body components}{that has front faces that can dominantly receive SRP on each body}.
Through the proposed methods, arbitrary transformable \Rep{solar sails}{spacecrafts} can acquire desired SRP acceleration without consuming any expendable propellant. These control methods are essential for zero-momentum transformable spacecrafts to accomplish desired orbit maneuver under solar radiation pressure.

\appendix
\subsection*{Appendix A}
Mathematical expressions in this appendix follow the formulations in Ref. \cite{attitude2007handbook}.
The explicit expressions in Equation \eqref{eq:gen_eom}, which can be derived through the Kane method \cite{kane1985dynamics}, are provided in the equations that follow.
Note that $f_X$ and $f^\ddag_X$ are external and inertial force exerted on the CoM of body $X$ respectively, whereas $n_X$ and $n^\ddag_X$ are external and inertial torque exerted on the \Del{center of mass of} body $X$ respectively. 
\Add{$n_{\theta_i}$ is the actuation torque of the $i$-th hinge joint. Other characters can be interpreted with the nomenclature list at the beginning of this paper.}
\begin{equation}\label{eq:gen_mass}
  \begin{split}
    &M_{vv} = \Rep{m}{m_\crm}U \qquad \\
    &M_{\omega\omega} = I_\crm\\
    &M_{\omega\theta, j}
    =\left( I_{\hat{j}}  -m_{\hat{j}} r_{\hat{j}\crm}^\times r_{\hat{j}\hrm_j}^\times \right)^\T \lambda_j\\
    &M_{\theta\omega, i}
    = \lambda_i^\T \left( I_{\hat{i}}  -m_{\hat{i}}  r_{\hat{i}\crm}^\times r_{\hat{i}\hrm_i}^\times\right)\\
    &M_{\theta\theta, ij} = M_{\theta\theta, ij}^*
    +\frac{m_{\hat{i}} m_{\hat{j}}}{\Rep{m}{m_\crm}} \lambda_i^\T \left( r_{\hat{i}\hrm_i}^\times r_{\hat{j}\hrm_j}^\times\right)^\T\lambda_j\\
    &M_{\theta \theta, ij}^*
    =
    \begin{cases}
        \lambda_i^\T\left( I_{\hat{j}} -m_{\hat{j}} r_{\hat{j}\hrm_i}^\times r_{\hat{j}\hrm_j}^\times\right)^\T\lambda_j &
        (j\in\hat{i}) \\
        \lambda_i^\T\left( I_{\hat{i}} +m_{\hat{i}}  r_{\hat{i}\hrm_i}^\times r_{\hat{i}\hrm_j}^\times\right)^\T \lambda_j &
        (i\in\hat{j}) \\
        0 & (i\notin\hat{j}\ \cap\ j\notin\hat{i})
    \end{cases}\\
    &M_{v\omega} = M_{v\theta} = M_{\omega v} = M_{\theta v} = O \qquad\text{(Zero matrix)}
  \end{split}
\end{equation}
\begin{equation}
  \begin{split}
    &d_{v}= \Rep{m}{m_\crm} \omega^\times\dot{R}_\crm\\
    &d_{\omega} = \sum_k\left\{\left(R_k-R_\crm\right)^\times f_k^\ddag+ n_k^\ddag\right\}\\
    &d_{\theta, i}
    = \lambda_i^\T\left[\sum_{k\in\hat{i}}\left\{\left(R_k-R_{\hrm_i}\right)^\times f_k^\ddag
    + n_k^\ddag\right\} -m_{\hat{i}}  r_{\hat{i}\hrm_i}^\times\omega^\times\dot{R}_\crm \right]
  \end{split}
\end{equation}
\begin{equation}
  \begin{split}
    &\tau_{v}
    = f_\crm\\
    &\tau_{\omega}
    = \sum_k\left\{\left(R_k-R_\crm\right)^\times f_k + n_k\right\}\\
    &\tau_{\theta, i}
    = \lambda_i^\T\left[\sum_{k\in\hat{i}}\left\{\left(R_k-R_{\hrm_i}\right)^\times f_k
    + n_k\right\}-\sum_k \frac{m_{\hat{i}} }{\Rep{m}{m_\crm}}\left( r_{\hat{i}\hrm_i}^\times f_k\right)\right]+n_{\theta_i}
  \end{split}
\end{equation}
where $M_{\omega\theta, j}$ is the $j$-th column of $M_{\omega\theta}$, $M_{\theta\omega, i}$ is the $i$-th row of $M_{\theta\omega}$, $d_{\theta, i}$ is the $i$-th component of $d_{\theta}$, and $\tau_{\theta, i}$ is the $i$-th component of $\tau_{\theta}$.

\subsection*{Acknowledgements}
\label{sec:ack}
The present study was supported by Advisory Committee for Space Engineering in Japan as a strategic research working group. The authors are grateful to the members of the Transformer working group that greatly enhanced the value of the present study through thorough and patient discussions.


\subsection*{Declaration of competing interest}
The authors declare that they have no known competing financial interests or personal relationships that could have appeared to influence the work reported in this paper.

\printbibliography


\end{document}